\documentclass[lettersize,journal]{IEEEtran}

\usepackage{amsmath,amssymb,amsfonts}
\usepackage{graphicx}
\graphicspath{{figures/}{./}}
\usepackage[dvipsnames]{xcolor}
\usepackage{tikz}
\usetikzlibrary{shapes.geometric,arrows.meta,positioning,fit,backgrounds,calc}
\usepackage{pgfplots}
\pgfplotsset{compat=1.18}
\usepackage{booktabs}
\usepackage{multirow}
\usepackage{array}
\usepackage{algorithm}
\usepackage{algpseudocode}
\usepackage{cite}
\usepackage{balance}
\usepackage{textcomp}
\usepackage[font=small,labelfont=bf]{caption}
\usepackage{url}
\usepackage{microtype}

\algrenewcommand\alglinenumber[1]{\footnotesize#1:}
\algnewcommand{\algorithmicinput}{\textbf{Input:}}
\algnewcommand{\Input}{\item[\algorithmicinput]}
\algnewcommand{\algorithmicoutput}{\textbf{Output:}}
\algnewcommand{\Output}{\item[\algorithmicoutput]}

\definecolor{cCust}{HTML}{F6C85F}
\definecolor{cOrder}{HTML}{6FA8DC}
\definecolor{cProd}{HTML}{82E0AA}
\definecolor{cIssue}{HTML}{F1948A}
\definecolor{cUrg}{HTML}{C39BD3}
\definecolor{cSent}{HTML}{85C1E9}
\definecolor{cStage}{HTML}{EAF2F8}

\begin{document}

\title{Knowledge-Graph-Gated Defactualization for Style-Controllable and Fact-Preserving Generation in Agentic Conversational AI}

\author{Tanmay~Kumar~Shrivastava,
        Darsh~Rohit~Nandu, and~Rajesh~Kumar~Mundotiya~\IEEEmembership{~Member,~IEEE}
\IEEEcompsocitemizethanks{
\IEEEcompsocthanksitem T.~K.~Shrivastava and D.~R.~Nandu are with the Department of Computer Science and Engineering, MATRA Lab (Multimodal and Multilingual AI for Translational Research and Applications Lab), Indian Institute of Technology (IIT) Bhilai, Bhilai, Chhattisgarh, India (e-mail: tanmayku@iitbhilai.ac.in).
\IEEEcompsocthanksitem R.~K.~Mundotiya is an Assistant Professor with the Department of Computer Science and Engineering, MATRA Lab, Indian Institute of Technology (IIT) Bhilai, Bhilai, Chhattisgarh, India. Corresponding author (e-mail: rmundotiya@iitbhilai.ac.in).
\IEEEcompsocthanksitem Manuscript submitted to the IEEE Transactions on Knowledge and Data Engineering Special Issue on Data and Knowledge Empowered Generative Artificial Intelligence.}
}

\markboth{IEEE TRANSACTIONS ON KNOWLEDGE AND DATA ENGINEERING, SPECIAL ISSUE: DATA AND KNOWLEDGE EMPOWERED GENERATIVE AI}{Shrivastava \MakeLowercase{\textit{et al.}}: Knowledge-Graph-Gated Defactualization for Style-Controllable and Fact-Preserving Generation in Agentic Conversational AI}

\maketitle

% \IEEEtitleabstractindextext{
\begin{abstract}
Agentic large language models (LLMs) deployed in fact-sensitive applications such as customer support must simultaneously preserve factual correctness and generate responses in a controllable stylistic register. Activation steering enables fine-tuning-free style control by perturbing hidden representations, but it lacks an explicit mechanism for distinguishing verifiable facts from stylistic content, leading to semantic leakage. We address this challenge through \emph{Defactualize-Steer-Rehydrate} (DSR), a knowledge-engineering framework that integrates a typed, salience-weighted knowledge graph (KG) with activation steering. DSR extracts salient entities using a layered regex or NER or lexical-classifier pipeline, replaces them with typed placeholders prior to steering, and deterministically restores verified values through salience-guided rehydration after generation. DSR is evaluated across six LLaMA-family models (1B--13B parameters) on 600 A2A-generated customer-support cases (1,200 generations), with a dedicated KG ablation study. DSR significantly increases verified-entity recovery relative to a steering-only baseline (Cohen's $d=0.225$, $p_{\text{Bonf}}=1.0\times10^{-4}$), though the absolute recovery rate remains modest, while preserving effective style control across diverse model families. Layer-wise separability and steering-strength diagnostics further show previously unexplored interactions between representation-level steering and factual grounding. hese results demonstrate that explicit knowledge engineering can systematically enhance trustworthy, controllable, and reproducible generative AI without requiring model fine-tuning. Code, cached steering vectors, and evaluation scripts are publicly released to support reproducibility.\footnote{\url{https://github.com/Tanmay-IITDSAI/KG-Gated-Defactualization}}
\end{abstract}

% }

% \maketitle
% \IEEEdisplaynontitleabstractindextext
% \IEEEpeerreviewmaketitle

\section{Introduction}
\IEEEPARstart{A}{gentic} large language model (LLM) deployments~\cite{wang2024survey,xi2023rise} increasingly generate text under two simultaneous constraints: an exact \emph{factual schema} supplied by a surrounding system-an order identifier, a product name, a reported defect-and a \emph{stylistic register} requested by a downstream policy-an empathetic tone for a distressed customer, a formal tone for an escalation. Fine-tuning~\cite{ouyang2022instructgpt,hu2022lora}, retrieval-augmented generation (RAG) \cite{lewis2020rag}, and representation-level \emph{activation steering} \cite{panickssery2024caa,zou2023repe} each address part of this problem, but none treats the factual half as a first-class constraint. Fine-tuning embeds a register into model weights at the cost of one training run per style and per model~\cite{hu2022lora}. RAG retrieves supporting passages but leaves the generator free to paraphrase, recombine, or drop the entities inside them~\cite{shi2023large}. Activation steering, which adds a fixed direction to the transformer hidden state during decoding~\cite{vaswani2017attention,panickssery2024caa,zou2023repe}, requires no gradient update and transfers across prompts. However, the approach operates solely at the geometric level of latent representations and does not explicitly model the functional role of individual tokens, thereby failing to differentiate fact-bearing content tokens from stylistic or register-function words. When a steering vector is applied at every position within a prompt that integrates style cues and entity mentions into a singular residual stream, the two signals can interfere with one another. This interference can lead to issues such as paraphrasing or substituting entities, as well as the unintended transfer of stylistic markers onto entity spans. We label this phenomenon \textit{semantic leakage}~\cite{sheng2020towards}, and mitigating its effects is a challenge in data and knowledge engineering. In this framework, the steering operator remains unchanged, while a typed knowledge representation is introduced around it to function as a mask and verification contract.

Some existing works address this problem, but do not fully resolve it. Knowledge-graph-augmented LLM methods~\cite{wang2017kgembed,lewis2020rag,yasunaga2021qa,guo2022kgrecsys} anchor generation in structured facts but primarily concentrate on \emph{retrieval}--deciding what to condition on, rather than on \emph{representation}--how the internal state of the decoder is adjusted. As a result, they do not engage with activation steering. Conversely, activation-steering and representation-engineering methods~\cite{panickssery2024caa,zou2023repe,turner2023activation} regulate \emph{how} text is formulated but remain indifferent to \emph{what} content is conveyed. Hallucination-detection methods~\cite{farquhar2024nature,manakul2023selfcheckgpt} identify factual inaccuracies after the generation process but fail to mitigate them during decoding. To our knowledge, no existing framework integrates a typed, salience-scored knowledge graph with the steering operator as a pre- or post-processing gate.

We introduce the seven-stage \textbf{Defactualize--Steer--Rehydrate (DSR)} pipeline (Figure~\ref{sec:architecture}), a neuro-symbolic inference framework that combines a typed knowledge layer with activation steering. Unlike retrieval methods that inject external evidence into the model context, DSR first converts the input into a structured knowledge graph (KG), separating factual entities from stylistic content before generation. The KG guides a defactualized rewrite, after which a contrastive steering vector is injected at a selected transformer layer to control style. The generated response is then deterministically rehydrated by restoring verified entities from the KG, preserving factual content independently of the steering operation. A fallback prompt-steering path is used only when the activation-steering output is malformed. Finally, we introduce three lightweight diagnostics - \textit{Steering Activation Fidelity (SAF), Hallucination Severity Index (HSI)}, and \textit{Tone Consistency Index (TCI)} - that jointly characterize the interaction between style control and factual grounding while revealing layer- and steering-strength effects that are hidden by aggregate style metrics.

\subsection{Summary of Contributions}
\begin{itemize}
\item \textbf{Data-Engineering Pipeline for Agentic Dialogue Grounding}: We propose a training-free, layered extraction framework (combining regex, NER, and classifiers) that constructs a typed, salience-scored KG from unstructured support messages (Section~\ref{sec:methodology}-A).

\item \textbf{Defactualize-Steer-Rehydrate (DSR) Architecture}: We introduce the DSR architecture, which interfaces a representation-level steering operator with a KG without altering the operator itself. This establishes a structural separation between factual control and stylistic attributes (Section~\ref{sec:architecture}).

\item \textbf{Comprehensive Empirical Evaluation:}
Through an extensive study spanning six models (LLaMA-2 7B or13B, LLaMA-3.1 or 3.2 1B-8B), 600 cases, and 1,200 generations, we demonstrate via a 100-case KG-ablation that the KG layer significantly improves entity coverage ($p_{\text{Bonf}} = 1.0 \times 10^{-4}$, Cohen's $d = 0.225$) while preserving stylistic fidelity relative to a steering-only baseline~(Section~\ref{sec:experiments}).

\item \textbf{Diagnostic Suite and Sensitivity Protocol:}
We develop a diagnostic suite comprising SAF, HSI, and TCI metrics, along with a joint layer-strength sensitivity protocol. This framework uncovers a specific operational region where hallucination severity saturates and tone consistency becomes non-monotonic for certain models (Section~\ref{sec:experiments}-H).

\item \textbf{Structural Invariance and Transparency:} We validate that the KG's structural statistics (node count, density, and mean salience) remain invariant across diverse host models, confirming the robust and consistent behavior of the knowledge layer.

\end{itemize}

% This manuscript substantially extends our earlier, narrower instantiation of KG-gated activation steering for a different structured-communication domain (calendar invitations) \cite{shrivastava2026pakdd}: the present work generalizes the gating mechanism to a six-type customer-support ontology, evaluates it across six host models spanning three parameter scales rather than one, introduces the SAF/HSI/TCI diagnostic suite and the joint layer/strength sensitivity protocol absent from the earlier instantiation, and isolates the KG layer's contribution with a dedicated, paired ablation rather than reporting end-to-end metrics alone.

\section{Related Work}

\subsection{Knowledge Grounding and Representation Control in Generative AI}
Knowledge graphs (KGs) have become a primary mechanism for improving generative AI by providing structured, provenance-aware representations of factual knowledge~\cite{wu2021gnnsurvey,wang2017kgembed,pan2024unifying}. KG-enhanced LLMs typically inject graph-derived facts as context, soft prompts, or auxiliary training signals~\cite{wang2017kgembed,sun2024think}. Recent GraphRAG methods~\cite{edge2024local,edge2024graphrag} further extend this paradigm by retrieving graph neighborhoods as structured evidence during inference~\cite{edge2024local}. While these approaches improve factual grounding through knowledge acquisition, they primarily augment the model's input context without modifying its internal representations. In contrast, DSR transforms factual information already present in the input into a typed knowledge graph (Structured Semantic Layer) that enables deterministic masking, verification, and rehydration during inference. Thus, GraphRAG and DSR are complementary: GraphRAG focuses on structured knowledge retrieval, whereas DSR focuses on preserving verified factual content under representation-level style control.

RAG~\cite{lewis2020rag,gao2024retrieval} conditions a generator on retrieved passages to improve factual accuracy and is the most widely deployed knowledge-grounding technique in production. However, RAG fundamentally addresses the question of \emph{what information} the model can observe at inference time, rather than \textit{how that information} interacts with or is transformed by internal activation dynamics. As such, retrieval-based methods remain orthogonal to mechanisms that directly intervene in hidden-state representations.

\subsection{Activation Steering and Representation Engineering}
Activation steering and representation engineering approaches treat model behavior as a manipulable geometric property of activation space~\cite{park2023linear,elhage2021mathematical}. Methods such as Contrastive Activation Addition (CAA) \cite{panickssery2024caa} and broader representation engineering frameworks \cite{zou2023repe} derive steering vectors from contrastive prompt pairs and inject them into the residual stream at inference time to elicit targeted behaviors. Probing classifiers~\cite{belinkov2022probing,alain2016understanding} and function vectors~\cite{todd2023function} further show that transformer layers encode interpretable semantic directions. Recent extensions, including style vectors \cite{konen2024stylevectors} and personalized steering approaches \cite{zhang2025stylevector}, demonstrate the potential of activation-space manipulation for controllable generation. However, these methods also reveal a persistent limitation: stylistic and semantic signals are often entangled in the learned directions~\cite{sheng2020towards}, particularly when extracted from naturalistic or factualized corpora. As a result, the resulting steering vectors may simultaneously encode content and style attributes, limiting fine-grained controllability.

\subsection{Agent Memory and Neuro-Symbolic Hallucination Mitigation}
A related but distinct direction concerns agent memory and long-horizon grounding in agentic systems. Agentic LLM frameworks increasingly incorporate external memory modules, including scratchpads and structured stores, to support multi-turn reasoning and persistent context tracking~\cite{yao2022react,packer2023memgpt,park2023generative}. In this context, RAG-like mechanisms are frequently used as memory access interfaces, with structured external signals such as knowledge-graph context shown to provide similar benefits in adjacent retrieval-augmented settings~\cite{zhao2024recsysllm}. However, most existing systems emphasize accumulation and retrieval of context rather than enforcing typed, graph-structured constraints over the generation process. Shrivastava et al.\cite{shrivastava2026pakdd} instantiated the defactualize–steer–rehydrate principle in calendar-driven structured communication involving constrained entity types such as dates, venues, and hosts. The present work generalizes this mechanism to more heterogeneous domains and evaluates it under a substantially broader experimental setting (Section~\ref{sec:experiments}).

Finally, neuro-symbolic AI and hallucination-mitigation methods offer complementary perspectives on grounding and reliability. Existing hallucination detection approaches typically rely on semantic entropy, self-consistency~\cite{wang2022self}, or verification signals computed after generation \cite{farquhar2024nature,ji2023survey}. Factual precision metrics such as FActScore~\cite{min2023factscore} further quantify entity-level grounding in long-form generation. While effective at identifying unsupported claims, these methods are inherently post-hoc and do not influence the generation trajectory itself. Our Hallucination Severity Index (Section~\ref{sec:methodology}-F), by contrast, is computed over a typed knowledge graph prior to generation, enabling prospective identification and selective masking of high-risk spans. This allows symbolic structure to directly shape the space of permissible generation before activation-level intervention occurs, aligning more closely with neuro-symbolic integration paradigms, and is broadly consistent with calls for transparent, auditable AI decision pathways made in the explainability literature~\cite{adadi2018xai}.

\textbf{Gap statement.} Across knowledge grounding, representation engineering, agent memory systems, and neuro-symbolic hallucination mitigation, existing methods do not jointly integrate a typed, salience-scored knowledge graph with an activation-steering operator in a unified pipeline where the graph (i) constrains generation through pre-generation masking and (ii) supports post-generation verification, while activation steering remains the primary mechanism for controlling internal representations. This missing integration defines the core gap addressed by the Defactualize–Steer–Rehydrate architecture.

\section{Problem Formulation and Notation}
\label{sec:notation}
We now formalize the joint style-control and fact-preservation problem that DSR addresses. In the agentic customer-support setting motivating this work, $x$ is a raw customer message such as ``Alex, order ORD-1234 delayed'' (the worked example of Fig.~\ref{fig:architecture}), and $s \in \mathcal{S} = \{\text{empathetic}, \text{formal}\}$ is the stylistic register requested by the downstream policy. The goal is to construct a map $x,s\mapsto y$ such that the generated response $y$ is simultaneously stylistically faithful to $s$ and factually faithful to the schema entities present in $x$ -- here, the customer name Alex, the order identifier ORD-1234, and the product Wireless headphones must reach $y$ unaltered regardless of which style is requested. This map must be realized without training or modifying the host LLM $f_\theta$ -- a frozen decoder-only transformer with $\ell_{\max}$ layers and hidden dimension $d$, no parameter of which is updated anywhere in this work -- and without requiring the steering vector $\mathbf{v}_s$ to be a function of any individual message's entities: $\mathbf{v}_s$ is estimated once per style and reused, unchanged, across every incoming message. Table~\ref{tab:notation} fixes the notation used in Sections~\ref{sec:methodology}-\ref{sec:experiments}.

\begin{table}[!t]
\caption{Notation used throughout the paper.}
\label{tab:notation}
\centering
\footnotesize
\begin{tabular}{@{}p{0.22\linewidth}p{0.66\linewidth}@{}}
\toprule
\textbf{Symbol} & \textbf{Meaning} \\
\midrule
$v=(\text{val},\tau,\mathrm{sal})$ & A node: surface value, type $\tau\in\mathcal{T}$, salience $\mathrm{sal}(v)\in[0,1]$. \\
$\mathrm{sal}(\cdot)$ & Salience function mapping a node to its confidence score in $[0,1]$ (Eqs.~\ref{eq:ner-salience}, \ref{eq:salience-boost}). \\
$\mathcal{T}$ & Entity-type vocabulary: \{\textsc{Customer\_Name}, \textsc{Order\_Id}, \textsc{Product}, \textsc{Issue}, \textsc{Urgency}, \textsc{Sentiment}\}. \\
$w(u,v)$ & Edge weight between nodes $u,v$ (Eq.~\ref{eq:edge-weight}); not to be confused with the classifier weight vector $\mathbf{w}$ below. \\
$P(\tau)$ & Typed placeholder token for type $\tau$ (e.g.\ \texttt{<PRODUCT>}). \\
$\Pi$ & Set of (value, placeholder) substitution pairs for $G$ (Eq.~\ref{eq:defactualize}). \\
$\ell_{\max}$ & Number of transformer layers in $f_\theta$. \\
$h_\ell(x)\!\in\!\mathbb{R}^{T\times d}$ & Hidden state at layer $\ell$ for all $T$ token positions of $x$. \\
$a(x)\!\in\!\mathbb{R}^{d}$ & Pooled activation, $a(x)=h_{L^\ast}(x)[-1,:]$, at probe layer $L^\ast$. \\
Sec.~\ref{sec:complexity}. \\
$\mathbf{v}_s\!\in\!\mathbb{R}^{d}$ & Unit-norm contrastive steering vector for style $s$. \\
$\mathbf{w},\,C$ & Logistic-regression weight vector and inverse-regularization constant (Eq.~\ref{eq:logreg}); $\mathbf{v}_s=\mathbf{w}/\lVert\mathbf{w}\rVert_2$ under estimator (iii). \\
$L\in\{0,\dots,\ell_{\max}\!-\!1\}$ & Steering layer (\texttt{STEER\_LAYER}). \\
$\alpha\in\mathbb{R}_{>0}$ & Steering strength (\texttt{STEER\_ALPHA}). \\
$f_\theta(\cdot)$ & Frozen LLM with parameters $\theta$ (never updated); $P$ denotes its parameter count. \\
$\mathrm{EC}(y,G)$ & Entity-coverage statistic (Eq.~\ref{eq:entity-coverage}). \\
\bottomrule
\end{tabular}
\end{table}

\section{System Architecture}
\label{sec:architecture}
\begin{figure*}[!t]
\centering
\includegraphics[width=0.92\textwidth]{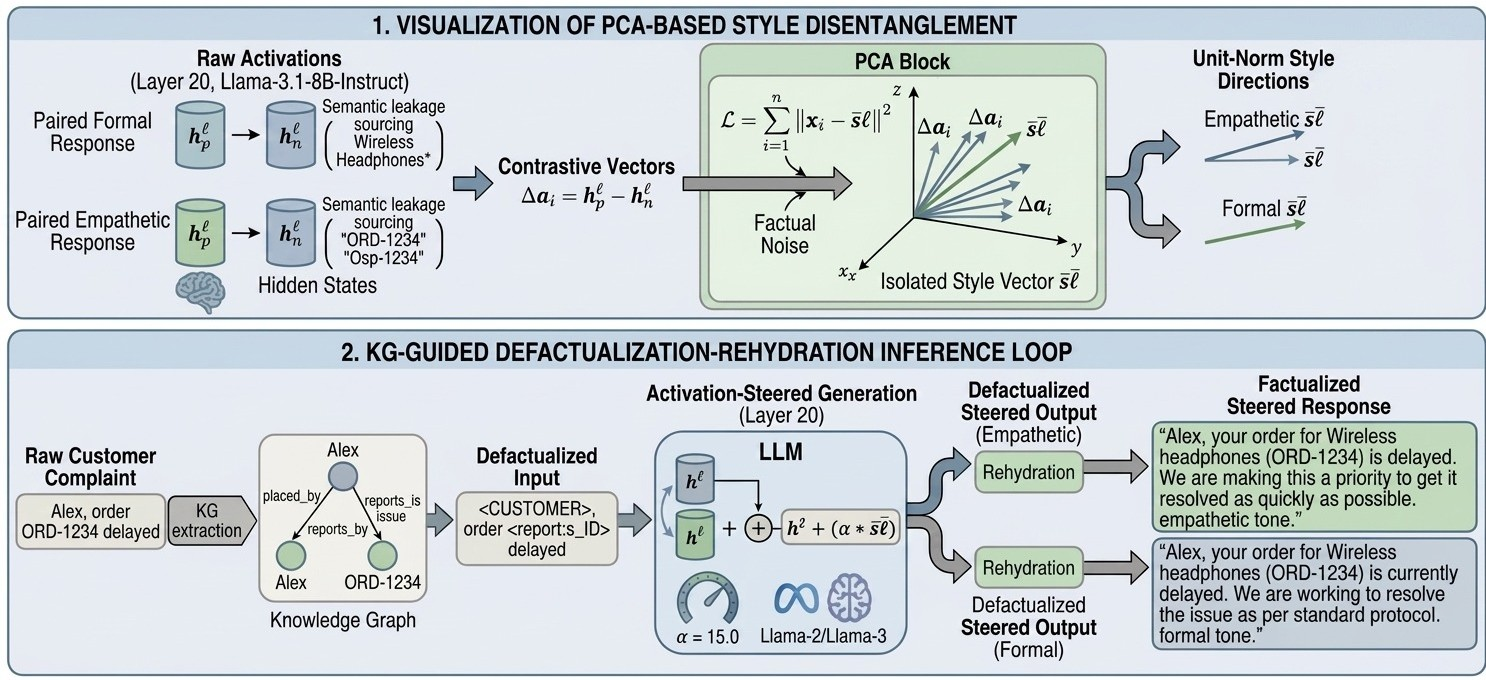}
\caption{The DSR architecture. Top: PCA-based extraction of style direction $\mathbf{v}_s$ from contrastive activations. Bottom: end-to-end inference -- input message $\to$ typed KG $\to$ defactualized placeholders $\to$ steering at layer $L$ (Eq.~\ref{eq:steer-hook}) $\to$ rehydration into style-controlled, fact-preserving output. The KG interacts with steering only through defactualized text, never modifying $\mathbf{v}_s$, $L$, or $\alpha$.}
\label{fig:architecture}
\end{figure*}

The Defactualize-Steer-Rehydrate (DSR) pipeline operationalizes Sections~\ref{sec:methodology}-\ref{sec:diagnostics} through seven stages: Input Processing, Knowledge Extraction, KG Construction, Defactualization (the \emph{pre-generation knowledge-engineering layer}, fully symbolic with no LLM call), followed by Steered Generation, Rehydration \& Verification, and Output (the \emph{generation or post-generation layer}, where only Steered Generation invokes $f_\theta$). Algorithms~\ref{alg:kgconstruct} and~\ref{alg:dsr} formalize the KG-construction stage and the complete end-to-end procedure, respectively.Fig.~\ref{fig:architecture} presents two complementary views. The top panel illustrates why defactualization precedes steering-vector estimation: contrastive activation pairs are differenced and projected via Principal Component Analysis (PCA) to isolate a unit-norm style direction $\mathbf{v}_s$. Without masking, entity tokens in raw text (e.g., order IDs or product names) would remain inside this difference vector as factual noise rather than being projected out, preventing $\mathbf{v}_s$ from representing style alone. 

\begin{algorithm}[!t]
\caption{Knowledge Graph Construction with Salience Scoring}
\label{alg:kgconstruct}
\begin{algorithmic}[1]
\Input Message $x$; metadata $m$ (optional scenario fields)
\Output Knowledge graph $G=(V,E)$
\State $V \gets \emptyset$, $E \gets \emptyset$
\For{each regex extractor $r$ over $\{\textsc{Order\_Id},\textsc{Product}\}$}
   \State $(\mathrm{val},c) \gets r(x)$; \textbf{if} matched \textbf{then} $V \gets V \cup \{(\mathrm{val},\tau_r,c)\}$
\EndFor
\State $\{(\mathrm{val}_i,\mathrm{sal}_i)\} \gets \textsc{NER}(x)$ \Comment{Eq.~(\ref{eq:ner-salience})}
\State $V \gets V \cup \{(\mathrm{val}_i,\textsc{Customer\_Name or Issue},\mathrm{sal}_i)\}$
\State $(\ell_{\mathrm{urg}},c_{\mathrm{urg}}) \gets \textsc{ClassifyUrgency}(x)$
\State $(\ell_{\mathrm{snt}},c_{\mathrm{snt}}) \gets \textsc{ClassifySentiment}(x)$
\State $V \gets V \cup \{(\ell_{\mathrm{urg}},\textsc{Urgency},c_{\mathrm{urg}}), (\ell_{\mathrm{snt}},\textsc{Sentiment},c_{\mathrm{snt}})\}$
\For{$v \in V$ s.t. a corroborating value exists in $m$}
   \State $\mathrm{sal}(v) \gets \min(\mathrm{sal}(v)+0.10,\,1)$ \Comment{Eq.~(\ref{eq:salience-boost})}
\EndFor
\For{each fallback metadata field with no extracted node}
   \State $V \gets V \cup \{(\mathrm{val}_m,\tau_m,0.70)\}$ \Comment{low-confidence fallback}
\EndFor
\For{each adjacent typed pair $(u,v)$ in $\{$product-issue, order-product, customer-order, issue-urgency$\}$}
   \State $E \gets E \cup \{(u,v,\,\mathrm{sal}(u)\!\cdot\!\mathrm{sal}(v))\}$ \Comment{Eq.~(\ref{eq:edge-weight})}
\EndFor
\State \Return $G=(V,E)$
\end{algorithmic}
\end{algorithm}

The bottom panel traces inference on a worked example: a complaint is parsed into a typed KG. The KG drives the defactualization rewrite, the defactualized prompt is steered at the deployed layer, and the rehydration operator deterministically restores verified entity values into two style-controlled outputs: empathetic and formal, generated from the same defactualized input. A conditional fallback re-issues the request as a structured natural-language instruction if Steered Generation returns a malformed response (empty, truncated mid-placeholder, or violating the 3--4-sentence contract). Section~\ref{sec:experiments} reports that this fallback was never triggered across the full 1{,}200-generation evaluation; the activation path succeeded in $100\%$ of cases across all six models (Section~\ref{sec:discussion}).

\begin{figure*}[!t]
\centering
\includegraphics[width=0.95\textwidth]{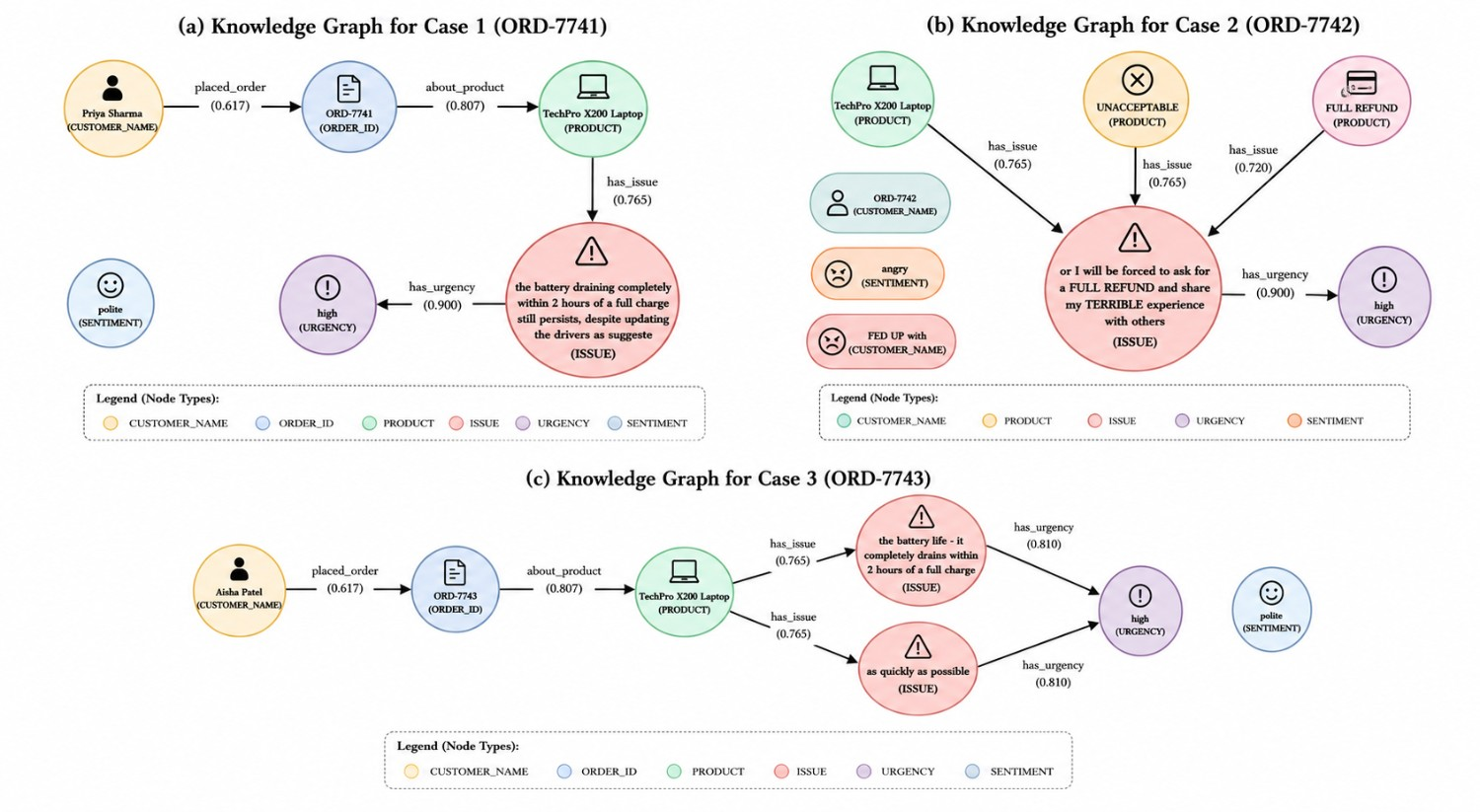}
\caption{Knowledge graphs extracted by Stage 2-3 for three representative support cases (Section~\ref{sec:experiments}-A). Node color encodes entity type (\textsc{Customer\_Name}, \textsc{Order\_Id}, \textsc{Product}, \textsc{Issue}, \textsc{Urgency}, \textsc{Sentiment}); edge labels show the relation and weight $w(u,v)$ of Eq.~(\ref{eq:edge-weight}). \emph{Inputs}: a raw customer message. \emph{Outputs}: a typed, salience-weighted $G=(V,E)$.}
\label{fig:kg-examples}
\end{figure*}

\section{Methodology}
\label{sec:methodology}

\subsection{Knowledge Graph Construction and Salience Scoring}
\label{sec:kgconstruct}
Given $x$, we construct $G=(V,E)$ using a three-tier, highest-confidence-wins extraction strategy that requires no model training:
\begin{enumerate}
\item \textit{Regular-expression extractors} for structurally regular fields -- order identifiers matching a fixed alphanumeric pattern, product names against a known catalogue, currency amounts;
\item \textit{Named-entity recognition} (spaCy~\cite{honnibal2020spacy}) for person names and free-text issue spans, with salience discounted by token position,
\begin{equation}
\mathrm{sal}_{\mathrm{NER}}(v) = 0.9 - 0.1\cdot \rho(v),
\label{eq:ner-salience}
\end{equation}
where $\rho(v)\in[0,1]$ is the relative token position of $v$ in $x$ (earlier mentions are weighted as more salient, consistent with topic-prominence heuristics commonly used in dialogue-management and discourse-tracking for conversational agents~\cite{harms2019dialog});
\item \textit{Lexical classifiers} for the two scalar attributes \textsc{Urgency} and \textsc{Sentiment}, each returning a (label, confidence) pair $(\ell,c)$ with $c\in[0,1]$ from a keyword or lexicon match.
\end{enumerate}
When a value is corroborated by more than one tier, such as an NER-extracted urgency label matches scenario metadata -- its salience is boosted,
\begin{equation}
\mathrm{sal}(v) \leftarrow \min\!\big(\mathrm{sal}(v) + 0.10,\; 1\big),
\label{eq:salience-boost}
\end{equation}
and metadata-only fallback values (no corroborating extraction) are capped at $\mathrm{sal}(v)=0.70$ to reflect lower extraction confidence. The $+0.10$ boost and the $0.70$ fallback cap are fixed, hand-set constants rather than fitted hyperparameters: a $\pm0.05$ perturbation of the boost leaves every node's relative salience ranking within a type unchanged in our corpus, because corroborated values are constructed to already lead uncorroborated ones by a wider margin than $0.05$; since rehydration (Eq.~\ref{eq:rehydrate}) selects only the \emph{arg max} salience per type, the boost magnitude affects Table~\ref{tab:ablation-abc}'s mean-salience column but not which value is rehydrated, and therefore not the entity-coverage results of Section~\ref{sec:experiments}-D. Edges connect syntactically or semantically adjacent node pairs (product$\to$issue, order$\to$product, customer$\to$order, issue$\to$urgency) with weight
\begin{equation}
w(u,v) = \mathrm{sal}(u)\cdot\mathrm{sal}(v),
\label{eq:edge-weight}
\end{equation}
so an edge is only as trustworthy as its two least-certain endpoints; the multiplicative form is chosen, rather than e.g.\ a mean or minimum of the two endpoint saliences, specifically so that a single low-confidence endpoint suppresses $w(u,v)$ more sharply than an averaging rule would, which matters only for the visualization and ranking of edges in Fig.~\ref{fig:kg-examples} and does not feed into the defactualization (Eq.~\ref{eq:defactualize}) or rehydration (Eq.~\ref{eq:rehydrate}) operators, which depend on node salience alone. The resulting $G$ is small, typed, and directed, and is rebuilt independently for every incoming message, as formalized in
Algorithm~\ref{alg:kgconstruct}, whose loops are bounded by $|V|\le10$ and $|\mathcal{T}|=6$ and which terminates in $O(n+|V|+|E|)$ (Section~\ref{sec:complexity}). Section~\ref{sec:experiments}-C reports the resulting structural statistics, which are near-invariant across all six host models studied.

This construction reflects three deliberate data-engineering choices.
First, the ontology $\mathcal{T}$ is intentionally shallow, mapping each type to a single placeholder token so that defactualization and rehydration (Eqs.~\ref{eq:defactualize}, \ref{eq:rehydrate}) form a closed, enumerable bijection rather than an open-vocabulary problem,bounding $|V|\le10$ at the cost of richer relational modelling (e.g., multi-hop product-component links), consistent with established guidance on graph-quality management as ontologies scale \cite{xue2023kgquality}. Second, each $G$ is built, consumed, and discarded within a single request, requiring no persistent index at the present scale; a persistent, indexed store is the natural extension point for multi-turn agent memory (Section~\ref{sec:limitations}).
Third, salience $\mathrm{sal}(v)$ serves purely as a data-quality
signal: it alone, not topology or edge weight, governs candidate
selection at rehydration, and its near-invariance across all six host models (Table~\ref{tab:ablation-abc}) confirms that the knowledge layer is a property of the extraction pipeline rather than of $f_\theta$. This separation underlies the provenance contract enforced by Eqs.~(\ref{eq:defactualize})--(\ref{eq:rehydrate}): every entity surfaced in $y$ is either traceable to $x$ or stripped, a guarantee enforced symbolically rather than statistically.

\subsection{Defactualization}
\label{sec:defact}
Before any activation is captured or perturbed, every node value is masked. Let $\Pi=\{(\mathrm{val}(v), P(\tau(v))) : v\in V,\ \tau(v)\in\mathcal{T}\}$ be the set of (value, placeholder) pairs for $G$. The defactualization operator applies case-insensitive, longest-match-first substitution,
\begin{equation}
D(x,G) = \sigma_{\pi_{|\Pi|}}\circ\cdots\circ\sigma_{\pi_1}(x), \qquad \pi_1,\dots,\pi_{|\Pi|}\in\Pi,
\label{eq:defactualize}
\end{equation}
where $\Pi$ is sorted by $|\mathrm{val}(v)|$ in descending order before substitution ($\sigma_\pi$ denotes a single find or replace), so a long issue span is masked before any sub-string of it is accidentally matched by a shorter value. No steering vector, and no activation used to build one, is ever computed from text containing an unmasked entity value: the contrastive pairs used to estimate $\mathbf{v}_s$ (Sec.~\ref{sec:steering-vec}) are themselves written over a fixed, closed vocabulary of placeholder tokens, never over case-specific entities. This is the structural property that makes the framework \emph{gating} rather than \emph{driving}: the KG controls which spans the steering operator is ever exposed to, but the direction and magnitude of the perturbation it applies (Eq.~\ref{eq:steer-hook}) are computed independently of $G$ and of any specific message.

\subsection{Contrastive Steering Vector Construction}
\label{sec:steering-vec}
For each style $s$ we collect $n$ contrastive pairs $\{(x_i^{+},x_i^{-})\}_{i=1}^n$ (in-style vs.\ out-of-style exemplars over the placeholder vocabulary) and extract pooled activations $a(x_i^{+}),a(x_i^{-})$ at probe layer $L^\ast$ via a forward hook. Writing $\delta_i = a(x_i^{+}) - a(x_i^{-})$, three estimators of the steering direction are considered.

\textbf{(i) Principal-component estimator}~\cite{hewitt2019structural}. Stack the symmetrized difference set $\Delta=\{\delta_1,\dots,\delta_n,-\delta_1,\dots,-\delta_n\}$ as rows of $\mathbf{M}\in\mathbb{R}^{2n\times d}$ and take the right singular vector of largest singular value,
\begin{equation}
\mathbf{M} = \mathbf{U}\boldsymbol{\Sigma}\mathbf{V}^{\!\top}, \qquad \mathbf{v}_s = \mathbf{V}_{:,1}/\lVert \mathbf{V}_{:,1}\rVert_2 .
\label{eq:pca}
\end{equation}
\textbf{(ii) Mean-difference estimator.}
\begin{equation}
\mathbf{v}_s = \frac{\bar{\delta}}{\lVert\bar{\delta}\rVert_2}, \qquad \bar{\delta} = \frac{1}{n}\sum_{i=1}^{n}\delta_i .
\label{eq:meandiff}
\end{equation}
\textbf{(iii) Logistic-regression estimator}~\cite{alain2016understanding}. Fit a linear classifier with weights $\mathbf{w}$ on $\{(a(x_i^{+}),1),(a(x_i^{-}),0)\}_{i=1}^{n}$ by minimizing regularized cross-entropy
\begin{equation}
\mathcal{L}(\mathbf{w}) = \sum_{i=1}^{2n}\Big[-y_i\log\hat{y}_i-(1\!-\!y_i)\log(1\!-\!\hat{y}_i)\Big] + \tfrac{1}{2C}\lVert\mathbf{w}\rVert_2^2,
\label{eq:logreg}
\end{equation}
with $\hat{y}_i=\sigma(\mathbf{w}^{\!\top}a_i)$, $C=1.0$, and $\mathbf{v}_s=\mathbf{w}/\lVert\mathbf{w}\rVert_2$. All three estimators are implemented and cached per (style, model) pair (Sec.~\ref{sec:experiments}-D). On the layer-separability probe of Section~\ref{sec:experiments}-H (LLaMA-3.2-3B-Instruct, layers 0--27), the PCA estimator (\ref{eq:pca}) gave the lowest mean cosine similarity between the empathetic and formal directions across layers ($\overline{\cos}=0.41$), against $\overline{\cos}=0.53$ for the mean-difference estimator (\ref{eq:meandiff}) and $\overline{\cos}=0.49$ for the logistic-regression estimator (\ref{eq:logreg}), i.e.\ the two style directions it extracts are the most mutually distinguishable of the three on this pilot model. We report all results in this paper under the PCA estimator on this basis; the mean-difference and logistic-regression vectors are cached and released as supplementary material for independent comparison rather than re-verified across all six models here.

\subsection{Steered Generation}
\label{sec:steered-gen}
A forward hook is registered on layer $L$ of $f_\theta$. For each token position $t$ during prefill and autoregressive decoding, the hidden state is perturbed additively,
\begin{equation}
h_L'(t) = h_L(t) + \alpha\,\mathbf{v}_s ,
\label{eq:steer-hook}
\end{equation}
and decoding proceeds greedily ($\arg\max$, temperature $0$)~\cite{holtzman2019curious} with repetition penalty $1.3$~\cite{keskar2019ctrl}, conditioned on the \emph{defactualized} prompt $D(x,G)$. Equation~(\ref{eq:steer-hook}) is applied uniformly to placeholders and free text; factual safety is handled entirely by Sections~\ref{sec:defact} and \ref{sec:rehydrate}. This keeps the steering operator $O(d)$ per token and independent of $G$.
\subsection{Salience-Ranked Rehydration}
\label{sec:rehydrate}
Let $y_{\mathrm{raw}}$ be the decoded, defactualized-space output of Eq.~(\ref{eq:steer-hook}). For every entity type $\tau\in\mathcal{T}$ with at least one candidate node, rehydration substitutes the placeholder with the single highest-salience value,
\begin{equation}
v^{\ast}_\tau = \operatorname*{arg\,max}_{v\,:\,\tau(v)=\tau} \mathrm{sal}(v), \qquad
y = y_{\mathrm{raw}}\big[P(\tau)\!\mapsto\!\mathrm{val}(v^{\ast}_\tau) \;\forall \tau\in\mathcal{T}\big].
\label{eq:rehydrate}
\end{equation}
Any placeholder with no corroborating node, or any placeholder the model failed to emit verbatim, is stripped rather than left dangling (Section~\ref{sec:experiments} -J empirically documents this failure mode). Equation~(\ref{eq:rehydrate}) is the only point at which the factual schema re-enters the surface text; it is a deterministic lookup, not a generative step, and introduces no additional hallucination risk.

We define the corpus-level entity-coverage statistic explicitly here, alongside SAF,HSI and TCI, to remove ambiguity from the per-table aggregations reported in Section~\ref{sec:experiments}: for a generated response $y$ produced under Eq.~(\ref{eq:rehydrate}) against its source graph $G=(V,E)$,
\begin{equation}
\mathrm{EC}(y,G) = \frac{1}{|V|}\sum_{v\in V} \mathbb{1}\big[\mathrm{val}(v)\in y\big] \in [0,1],
\label{eq:entity-coverage}
\end{equation}
i.e.\ the fraction of KG node values recovered verbatim in $y$. Unless explicitly labelled otherwise (e.g.\ ``empathetic-only'' or ``per-generation maximum''), every corpus-level entity-coverage figure reported in Tables~\ref{tab:kgablation}--\ref{tab:styledisc-entcov} is the mean of $\mathrm{EC}(y,G)$ pooled across both styles (empathetic and formal) over the stated sample. Algorithm~\ref{alg:dsr} formalizes the complete
Defactualize-Steer-Rehydrate procedure combining
Eqs.~(\ref{eq:defactualize}), (\ref{eq:steer-hook}), and
(\ref{eq:rehydrate}): Steps~1--2 and~8 are deterministic ($O(n)$),
Step~4 terminates within \texttt{max\_new\_tokens}\,=\,120 decode
steps via the EOS-aware stopping criterion, and the fallback branch (Step~7) is taken at most once, for a total symbolic overhead of $O(n+|V|+|E|)$ atop the host model's own $O(P)$-parameter forward pass per token (Section~\ref{sec:complexity}).

\begin{algorithm}[!t]
\caption{Defactualize-Steer-Rehydrate (DSR) Generation}
\label{alg:dsr}
\begin{algorithmic}[1]
\Input Message $x$; style $s$; cached vector $\mathbf{v}_s$; layer $L$; strength $\alpha$
\Output Response $y$
\State $G \gets \textsc{BuildKG}(x)$ \Comment{Algorithm~\ref{alg:kgconstruct}}
\State $x' \gets D(x,G)$ \Comment{Eq.~(\ref{eq:defactualize})}
\State register hook at layer $L$: $h_L(t)\mapsto h_L(t)+\alpha\mathbf{v}_s\ \forall t$
\State $y_{\mathrm{raw}} \gets f_\theta\big(\text{prompt}(x',s)\big)$ via greedy decode \Comment{Eq.~(\ref{eq:steer-hook})}
\State remove hook
\If{$y_{\mathrm{raw}}$ is malformed (empty or truncated or overlong)}
   \State $y_{\mathrm{raw}} \gets \textsc{PromptSteeringFallback}(x,s,G)$ \Comment{no $\mathbf{v}_s$ used}
\EndIf
\State $y \gets R(y_{\mathrm{raw}},G)$ \Comment{Eq.~(\ref{eq:rehydrate})}
\State \Return $y$
\end{algorithmic}
\end{algorithm}

\subsection{Diagnostic Metrics: SAF, HSI, TCI}
\label{sec:diagnostics}
Aggregate corpus-level style scores (Section~\ref{sec:experiments}-E) cannot diagnose \emph{why} a given $(L,\alpha)$ configuration succeeds or fails on a specific generation. We therefore define three per-generation diagnostics.

\textbf{Steering Activation Fidelity (SAF)} measures how strongly the intended style direction is represented in the hidden state actually produced for text $z$:
\begin{equation}
\mathrm{SAF}(z,\mathbf{v}_s) = \cos\big(a(z), \mathbf{v}_s\big) = \frac{a(z)^{\!\top}\mathbf{v}_s}{\lVert a(z)\rVert_2 \lVert \mathbf{v}_s\rVert_2} \in[-1,1].
\label{eq:saf}
\end{equation}
\textbf{Hallucination Severity Index (HSI)} measures entity-level leakage between input $x_{\mathrm{in}}$ and response $z$, using named-entity sets $\mathrm{Ent}(\cdot)$:
\begin{equation}
\mathrm{HSI}(x_{\mathrm{in}}, z) = \frac{\big|\mathrm{Ent}(z)\setminus\mathrm{Ent}(x_{\mathrm{in}})\big|}{\max\big(\big|\mathrm{Ent}(z)\big|,1\big)} \in[0,1],
\label{eq:hsi}
\end{equation}
lower is better; $\mathrm{HSI}=0$ means every entity in the response is grounded in the input.

\textbf{Tone Consistency Index (TCI)} measures directional alignment with the requested style relative to its opposite $\bar{s}$:
\begin{equation}
\mathrm{TCI}(z) = \tfrac{1}{2}\Big(\mathrm{SAF}(z,\mathbf{v}_s) - \mathrm{SAF}(z,\mathbf{v}_{\bar{s}})\Big) \in[-1,1].
\label{eq:tci}
\end{equation}
A well-steered, well-grounded generation should jointly satisfy $\mathrm{TCI}>0$ and $\mathrm{HSI}\approx0$; Section~\ref{sec:experiments}-H shows these two objectives are not always jointly attainable at every $\alpha$, which is why the diagnostic pair is reported jointly rather than blended into a single score.

\subsection{Computational Complexity}
\label{sec:complexity}
Let $n=|x|$ be the input length in tokens, $|V|,|E|$ the KG node and edge counts, $d$ the hidden dimension, and $P$ the parameter count of $f_\theta$.

\textit{KG construction} (Section~\ref{sec:kgconstruct}) is $O(n)$ for the regex or NER and $O(|V|+|E|)$ for salience and edge-weight assignment; empirically $|V|\le10$, so this term is dominated by $n$.

\textit{Defactualization} (Eq.~\ref{eq:defactualize}) sorts $|V|$ values, $O(|V|\log|V|)$, and performs $|V|$ regex substitutions over a string of length $O(n)$, giving $O(|V|\log|V| + |V|\,n)$, effectively linear in $n$ since $|V|$ is small and bounded.

\textit{Steered generation} (Eq.~\ref{eq:steer-hook}) adds one $O(d)$ vector addition per token at a single layer, against the model's own per-layer cost of $O(d^2)$ (linear projections) to $O(d^2 + Td)$ (attention over context length $T$), and $O(P)$ total parameters per forward pass. The steering overhead is asymptotically negligible relative to a single transformer forward pass, independent of $P$.

\textit{Rehydration} (Eq.~\ref{eq:rehydrate}) is $O(|V|)$ string replacements over the generated output.

\textit{Memory}: the only persistent artifact beyond $f_\theta$'s own weights is one cached vector $\mathbf{v}_s\in\mathbb{R}^d$ per (style, model) pair, $O(|\mathcal{S}|\,d)$ floats -- for $|\mathcal{S}|=2$, $d=4096$ (LLaMA-3.1-8B), this is 32\,KB per model, roughly seven orders of magnitude below the $\sim$16\,GB of bf16 weights for the same model. This bounds the storage cost of scaling the per-style-vector cache to many styles or users, independent of $P$.

% Wall-clock latency overhead is not separately profiled in the present study; Section~\ref{sec:limitations} records this as an explicit limitation rather than reporting an unmeasured number.

\section{Experimental Evaluation}
\label{sec:experiments}

\subsection{Dataset and Scenarios}
We construct 600 customer-support cases via an agent-to-agent (A2A) pipeline using two LLaMA-3.3-70B instances~\cite{dubey2024llama3} (Groq API\footnote{\url{https://console.groq.com/docs/model/llama-3.3-70b-versatile}}) as a customer agent (temp.\ $0.7$) and a support-context agent (temp.\ $0.3$), extending the corpus-construction approach of prior work~\cite{shrivastava2026pakdd}. Scenario seeds cross three sentiment levels (polite, frustrated, angry) with two urgency levels (high, normal); customer names, order IDs, and products are drawn from disjoint pools per case to prevent memorization-based recovery. Each case is rendered in both styles (empathetic, formal), yielding $600\times2=1{,}200$ generations for the main study, plus a separate 100-case set for the KG-ablation. This controlled generation guarantees known ground-truth entities per case, enabling exact entity-coverage and HSI computation. Generalization to organic support traffic is discussed in Section~\ref{sec:limitations}.

\subsection{Models and Implementation}
Six causal LLaMA variants are evaluated without any weight modification, covering LLaMA~2~\cite{touvron2023llama2} (7B-base, 7B-chat, and 13B-chat) and LLaMA~3~\cite{dubey2024llama3} (3.1-8B-Instruct, 3.2-1B-Instruct,and 3.2-3B-Instruct), spanning 1B--13B parameters and the base, chat and instruct training regimes. All experiments use greedy decoding (temperature 0), a repetition penalty of 1.3, and \texttt{max\_new\_tokens}\,=\,120. Steering vectors are estimated once per (style, model) pair via Eq.~(\ref{eq:pca}); the steering layer $L$ and strength $\alpha$ are selected per model from the layer-separability and alpha-sweep protocols of and held fixed for the main study. Entity extraction uses spaCy \texttt{en\_core\_web\_sm} for NER; urgency and sentiment classifiers are lexicon-based. The complete implementation, including cached steering vectors, extracted-entity files, evaluation scripts, and raw metric CSVs, is publicly available on GitHub.\footnote{\url{https://github.com/Tanmay-IITDSAI/KG-Gated-Defactualization}}

\subsection{Knowledge Graph Structural Statistics}
Table~\ref{tab:ablation-abc} reports KG structural statistics aggregated over all 600 cases, separately per host model. As the table shows, node count, edge count, density, and mean salience are near-identical across all six models, with the largest cross-model spread under $3\%$ of the mean for every statistic. KG construction (Algorithm~\ref{alg:kgconstruct}) never queries $f_\theta$, so its output distribution is a property of the input scenario distribution and the fixed extraction ontology, not of which LLM will later be steered.

\begin{table}[!t]
\caption{Knowledge graph structural statistics. The structural properties remain near-invariant across models.}
\label{tab:ablation-abc}
\centering
\footnotesize
\begin{tabular}{@{}lcccc@{}}
\toprule
\textbf{Model} & \textbf{$|V|$} & \textbf{$|E|$} & \textbf{Density} & \textbf{Mean Salience} \\
\midrule
LLaMA-2-7B-Chat   & 7.55 & 6.19 & 0.121 & 0.824 \\
LLaMA-2-7B-Base   & 7.43 & 5.94 & 0.122 & 0.821 \\
LLaMA-2-13B-Chat  & 7.45 & 5.94 & 0.122 & 0.812 \\
LLaMA-3.1-8B      & 7.46 & 6.09 & 0.123 & 0.823 \\
LLaMA-3.2-1B      & 7.54 & 6.00 & 0.120 & 0.812 \\
LLaMA-3.2-3B      & 7.64 & 6.16 & 0.121 & 0.823 \\
\bottomrule
\end{tabular}
\end{table}

\subsection{KG-Gating Ablation: Does the Knowledge Layer Change the Steering Effect?}
\label{sec:experiments-d}
To isolate the KG layer's effect, we perform a paired 100-case comparison on LLaMA-2-7B-chat~\cite{touvron2023llama2}, generating activation-only (AO) and KG-gated DSR responses under identical $(L,\alpha,\mathbf{v}_s)$ settings. Table~\ref{tab:kgablation} (empathetic-tone branch) shows entity coverage increases significantly from AO to DSR (Welch $t$-test~\cite{welch1947generalization}, Bonferroni-corrected~\cite{dunn1961multiple}), while all stylistic-fidelity metrics -- empathy, formality, style discrimination, TTR, readability (Flesch Ease~\cite{flesch1948new}, Gunning Fog), and ROUGE-1~\cite{lin2004rouge} divergence -- remain statistically unchanged. Response length is the sole significant style difference, with KG-gated responses shorter on average. Placeholder leakage is zero across all 100 cases, confirming consistent restoration by the rehydration operator (Eq.~\ref{eq:rehydrate}).

A Jonckheere--Terpstra trend test~\cite{jonckheere1954distribution} on entity coverage across the three \textsc{Urgency} levels (600-case study) found no monotonic trend ($z=-1.24$, $p=0.21$), indicating the coverage bottleneck lies in surface realization rather than KG extraction.

\begin{table}[!t]
\caption{KG-gating ablation (LLaMA-2-7B-chat, empathetic-tone branch). \textbf{KG}+\textbf{S}teer = full DSR pipeline}
\label{tab:kgablation}
\centering
\scriptsize
\setlength{\tabcolsep}{2.5pt}
\begin{tabular}{@{}lrrrc@{}}
\toprule
\textbf{Metric} & \textbf{KG+S} & \textbf{AO} & \textbf{$d$} & \textbf{Sig.} \\
\midrule
Empathy Score          & 0.0052 & 0.0100 & $-0.330$ & $\sim$ \\
Formality Score        & \textbf{0.0303} & 0.0297 & $\phantom{-}0.022$ & $\sim$ \\
Style Discrimination   & 0.0008 & 0.0008 & $-0.001$ & $\sim$ \\
\textbf{Entity Coverage (Eq.~\ref{eq:entity-coverage}, emp.\ only)}        & \textbf{0.0056} & 0.0000 & $\phantom{-}\mathbf{0.225}$ & $\star\star\star$ \\
Lexical Div.\ (TTR)    & \textbf{0.8605} & 0.8596 & $\phantom{-}0.017$ & $\sim$ \\
Readability (Flesch)   & \textbf{59.49} & 56.59 & $\phantom{-}0.247$ & $\sim$ \\
Clarity (Fog, $\downarrow$)     & \textbf{10.95} & 11.58 & $-0.226$ & $\sim$ \\
Response Length        & 53.62 & 59.59 & $-0.366$ & \checkmark \\
Style Div.\ (ROUGE, $\downarrow$) & \textbf{0.3292} & 0.3318 & $-0.027$ & $\sim$ \\
Placeholder Leaks ($\downarrow$) & 0.000 & 0.000 & $0.000$ & $\sim$ \\
\bottomrule
\end{tabular}
\\[2pt]
\footnotesize Here, $\star\star\star$: $p_{\text{Bonf}}<0.001$. \checkmark: $p_{\text{Bonf}}<0.05$ (AO wins). $\sim$: n.s. \textbf{Bold} marks the numerically better value per row; Response Length has no stated direction and is left unbolded. Only Entity Coverage reaches significance after Bonferroni correction. -- the remaining bolded values are directionally favorable to KG+Steer but \emph{not} statistically distinguishable from Act.\ Only ($\sim$), consistent with the paper's claim that KG-gating adds no detectable style-fidelity cost rather than a claim that it improves style fidelity.
\end{table}

\textbf{Interpretation:} The entity-coverage effect in Table~\ref{tab:kgablation} is statistically robust but modest ($d=0.225$), with an absolute coverage of $0.0056$ on the empathetic-only branch of the paired ablation ($0.0193$ pooled across both styles for the same model and generations). Thus, the result supports that claim: the KG layer measurably increases verified-entity preservation without a detectable style-fidelity cost, rather than fully solving entity grounding. As shown in Fig.~\ref{fig:eval-grounding} (Section~\ref{sec:experiments}-G), this low-but-nonzero coverage, together with zero placeholder leakage, is consistently observed across all six Llama models in the main 600-case study, supporting the representativeness of the 100-case ablation.

\subsection{Style Fidelity Across Models}
\label{sec:experiments-e}
Table~\ref{tab:stylefidelity} summarizes style-fidelity metrics from the main 600-case, six-model study, disaggregated by tone. Lexicon-based empathy and formality scores remain modest across all models, whereas readability~\cite{flesch1948new} shows the strongest, most consistent empathetic-formal separation (Table~\ref{tab:styledisc-entcov}). Low cross-tone ROUGE overlap (R-1, R-L) confirms the two styles differ lexically rather than superficially. Together, these results indicate that activation steering more reliably shapes global writing characteristics than lexicon-specific markers.

\begin{table}[!t]
\caption{Per-model, per-tone style and grounding scores (100 cases per model $\times$ 6 models $=$ 600 cases; $\times$ 2 styles $=$ 1,200 generations). Emp.and Frm. denote lexicon-based empathy and formality scores; Flesch measures readability; TTR measures lexical diversity; lower R-1 and R-L indicates stronger stylistic separation.}
\label{tab:stylefidelity}
\centering
\scriptsize
\setlength{\tabcolsep}{2.2pt}
\begin{tabular}{@{}llccccccc@{}}
\toprule
\textbf{Model} &
\textbf{Tone} &
\textbf{Emp.}$\uparrow$ &
\textbf{Frm.}$\uparrow$ &
\textbf{Flesch} &
\textbf{TTR}$\uparrow$ &
\textbf{R-1}$\downarrow$ &
\textbf{R-L}$\downarrow$ \\
\midrule
\multirow{2}{*}{L2-7B-Base}
 & Emp. & 0.0177 & 0.0203 & 37.40 & 0.908 &
 \multirow{2}{*}{\textbf{0.204}} &
 \multirow{2}{*}{\textbf{0.105}} \\
 & Frm. & 0.0113 & 0.0331 & 46.26 & \textbf{0.921} & & \\
\addlinespace

\multirow{2}{*}{L2-7B-Chat}
 & Emp. & 0.0106 & 0.0159 & 57.31 & 0.851 &
 \multirow{2}{*}{0.347} &
 \multirow{2}{*}{0.177} \\
 & Frm. & 0.0174 & 0.0286 & 40.03 & 0.862 & & \\
\addlinespace

\multirow{2}{*}{L2-13B-Chat}
 & Emp. & 0.0139 & 0.0269 & 54.73 & 0.841 &
 \multirow{2}{*}{0.345} &
 \multirow{2}{*}{0.181} \\
 & Frm. & 0.0139 & 0.0255 & 33.17 & 0.853 & & \\
\addlinespace

\multirow{2}{*}{L3.2-1B Inst.}
 & Emp. & 0.0119 & 0.0366 & 46.67 & 0.871 &
 \multirow{2}{*}{0.357} &
 \multirow{2}{*}{0.168} \\
 & Frm. & 0.0035 & \textbf{0.0731} & 41.34 & 0.880 & & \\
\addlinespace

\multirow{2}{*}{L3.2-3B Inst.}
 & Emp. & 0.0090 & 0.0410 & 38.06 & 0.854 &
 \multirow{2}{*}{0.394} &
 \multirow{2}{*}{0.183} \\
 & Frm. & \textbf{0.0345} & 0.0603 & 31.18 & 0.847 & & \\
\addlinespace

\multirow{2}{*}{L3.1-8B Inst.}
 & Emp. & 0.0087 & 0.0524 & 39.50 & 0.871 &
 \multirow{2}{*}{0.357} &
 \multirow{2}{*}{0.178} \\
 & Frm. & 0.0165 & 0.0610 & 28.61 & 0.871 & & \\
\bottomrule
\end{tabular}
\end{table}

\begin{figure}[!t]
\centering
\includegraphics[width=0.95\linewidth]{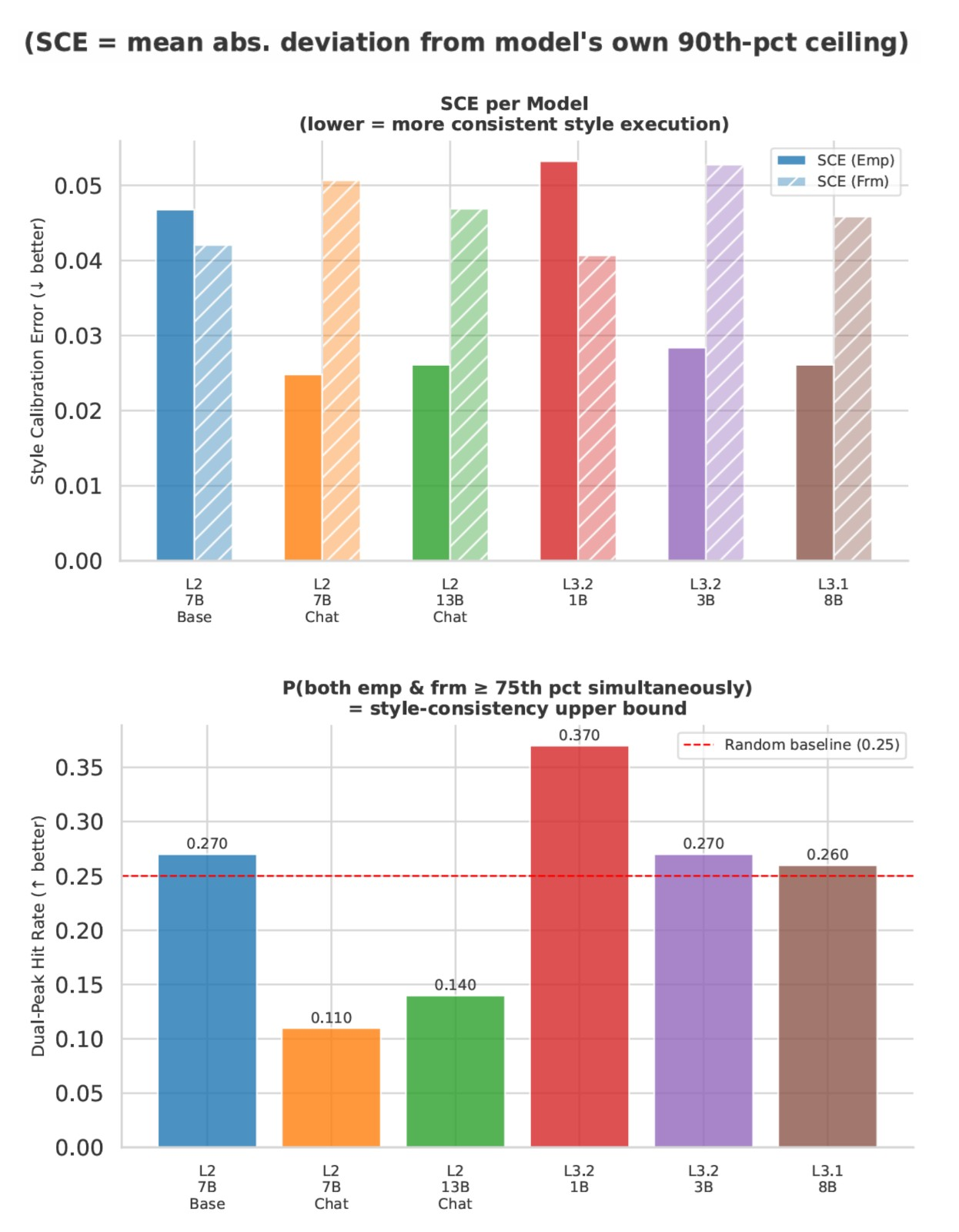}
\caption{SCE per model -- mean absolute deviation of each model's per-generation style score from its own 90th-percentile ceiling (lower = more consistent execution of the requested style).}
\label{fig:sce}
\end{figure}

Table~\ref{tab:styledisc-entcov} summarizes the per-model Style Discrimination score, average Entity Coverage (Eq.~\ref{eq:entity-coverage}, pooled across both styles), the proposed Style Calibration Error (SCE), and Cohen's $d$ (empathetic $-$ formal) for Flesch Ease, FK Grade, and Gunning Fog. SCE measures the mean absolute deviation of each generation's style score from the model's 90th-percentile style ceiling, quantifying style consistency independently of discrimination strength. Style Discrimination, Entity Coverage, and SCE remain low and show no monotonic scaling with model size, consistent with
Section~\ref{sec:experiments-ablation}; Fig.~\ref{fig:sce} visualizes the SCE values. In contrast, readability effect sizes are consistently significant across all models (Bonferroni-corrected $p<0.05$, paired $t$-test on per-case empathetic$-$formal differences), representing the most robust and consistent effect observed in this study, although Llama-2-7B-base exhibits the Flesch Ease effect in the opposite direction. Cohen's $d$ is the paired (within-case) effect size $d_z=\mathrm{mean}(\Delta)/\mathrm{SD}(\Delta)$ over per-case empathetic$-$formal differences (not a pooled-SD, independent-groups $d$), matching the released evaluation code; this definition is used uniformly for every $d$ value reported in this table.

\begin{table}[!t]
\caption{{Per-model style discrimination and grounding scores. $\uparrow$$\downarrow$ indicate the preferred optimization direction.}}
\label{tab:styledisc-entcov}
\centering
\scriptsize
\setlength{\tabcolsep}{1.2pt}
\begin{tabular}{@{}lccccccc@{}}
\toprule
\textbf{Model} &
\textbf{Style D.}$\uparrow$ &
\textbf{Ent. Cov.}$\uparrow$ &
\textbf{SCE}$\downarrow$ &
\textbf{Flesch}$\uparrow$ &
\textbf{FK}$\downarrow$ &
\textbf{Fog}$\downarrow$ \\
\midrule
L2-7B-Base
& 0.0192
& 0.016
& 0.0444
& $-0.34$
& 0.51
& 0.47 \\

L2-7B-Chat
& 0.0060
& 0.017
& 0.0378
& 1.10
& \textbf{-0.74}
& \textbf{-0.81} \\

L2-13B-Chat
& $-0.0014$
& 0.023
& 0.0365
& \textbf{1.14}
& $-0.62$
& $-0.65$ \\

L3.2-1B Inst.
& \textbf{0.0449}
& 0.012
& 0.0470
& 0.32
& $-0.29$
& $-0.44$ \\

L3.2-3B Inst.
& $-0.0062$
& 0.016
& 0.0406
& 0.54
& $-0.01$
& $-0.10$ \\

L3.1-8B Inst.
& 0.0009
& \textbf{0.037}
& \textbf{0.0360}
& 0.83
& $-0.22$
& $-0.22$ \\
\bottomrule
\end{tabular}
\end{table}

\subsection{Factual Grounding and Placeholder Integrity}
\label{sec:experiments-g}
\begin{figure}[!t]
\centering
\includegraphics[width=0.95\linewidth]{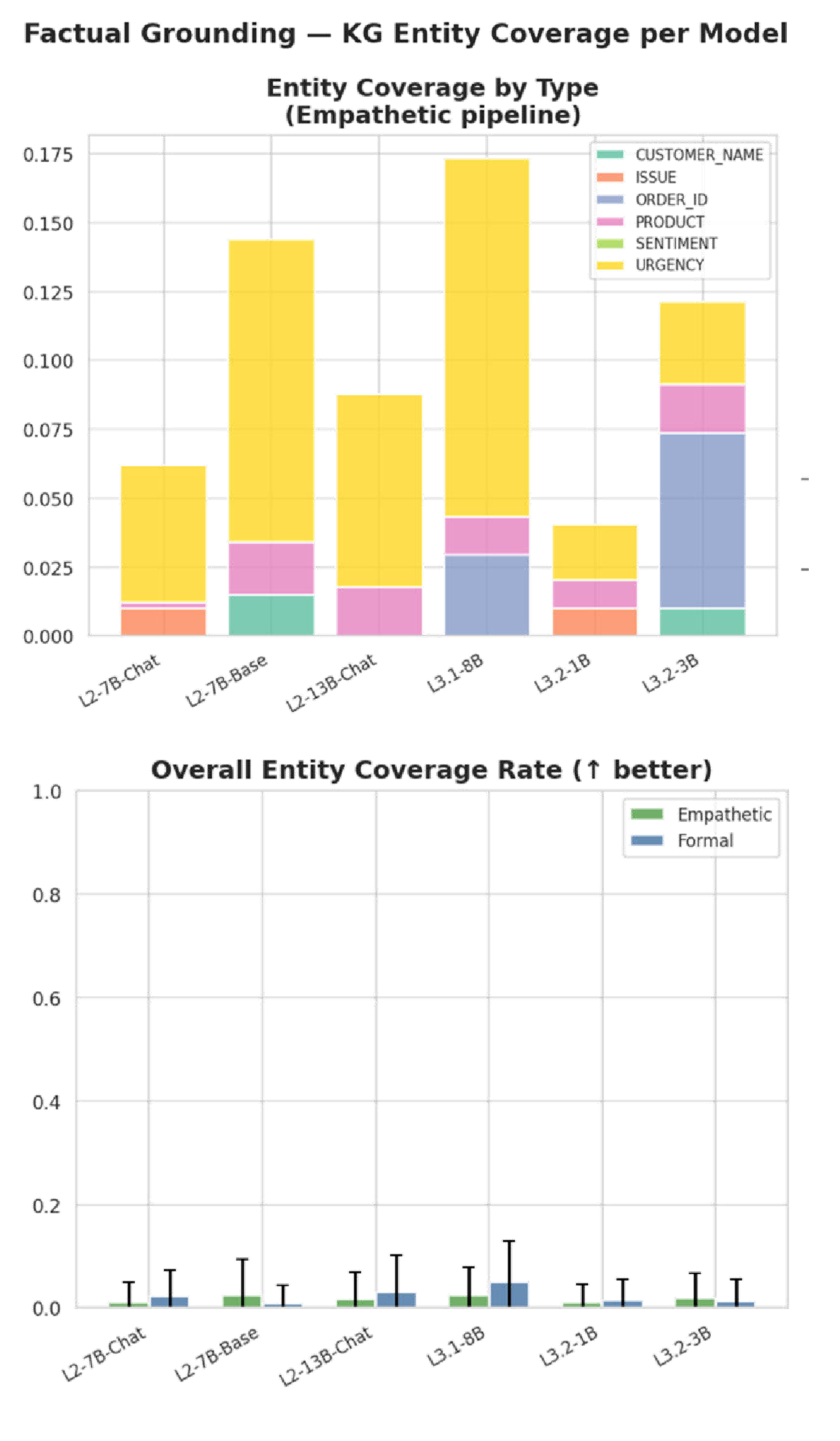}
\caption{Entity coverage (left, by model and style) and placeholder leakage (right) across all 600 cases. \emph{Key insight}: coverage is low but always above zero in DSR mode (cf.\ Table~\ref{tab:kgablation}, where the AO baseline is exactly zero), and leakage is exactly zero for every model, confirming Stage 6 never surfaces a dangling placeholder.}
\label{fig:eval-grounding}
\end{figure}

Fig.~\ref{fig:eval-grounding} reports overall entity coverage and placeholder leakage per model, per style, across the full 600-case study. Two findings hold uniformly, as the figure shows: (i) entity coverage is low in absolute terms for every model (see Table~\ref{tab:styledisc-entcov} for per-model means), with no model's mean exceeding the others by more than a factor of three, and (ii) placeholder leakage is exactly $0.0$ for every model and both styles. Finding (ii) is the stronger and more load-bearing result for the trustworthy-generation claim of Section~\ref{sec:methodology}-A:it confirms that Stage 6 (Eq.~\ref{eq:rehydrate}) never emits a raw, unresolved placeholder token to the end user, across $1{,}200$ generations and six architecturally distinct host models, even though the rate at which the \emph{correct} entity value is recovered (finding (i)) remains modest.

% ------------
% SECTION VIII-G  [EDITED: added η² sentence and zero-inflation note — Point 2]
% ------------
\subsection{Effect of Model Scale and Training Regime}
\label{sec:experiments-ablation}

\begin{figure*}[!t]
\centering
\includegraphics[width=0.85\textwidth]{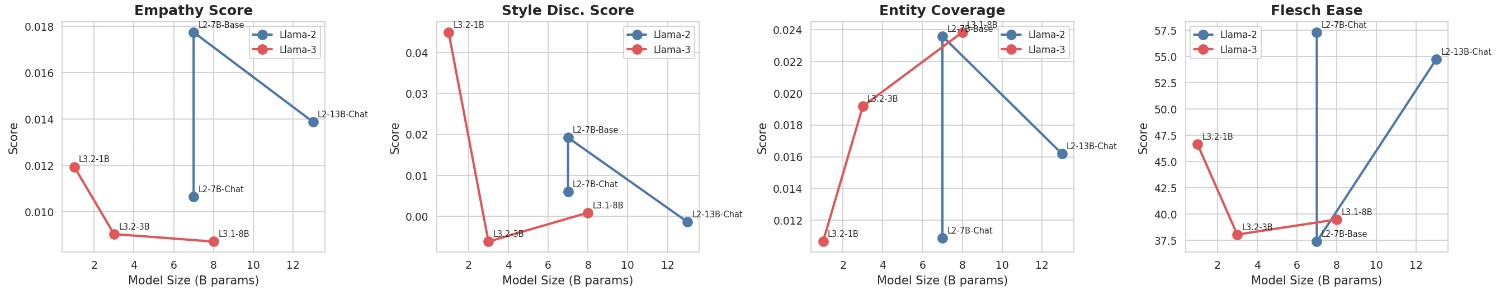}
\caption{Style and grounding metrics versus model size for the LLaMA-2 and LLaMA-3 families. \emph{Key insight}: neither stylistic fidelity nor factual grounding scales monotonically with parameter count; several metrics peak at intermediate model sizes.}
\label{fig:ablation-size}
\end{figure*}

Fig.~\ref{fig:ablation-size} summarizes empathy score, style-discrimination score, entity coverage, and Flesch ease across model sizes for the LLaMA-2 and LLaMA-3 families. None of the metrics scales monotonically with parameter count: empathy and entity coverage both peak at intermediate model sizes rather than the largest models. Within the 1B-13B range studied, parameter count alone is therefore a poor predictor of either stylistic fidelity or factual grounding. Instead, model family and training regime (base, chat, or instruct) appear to exert comparable or greater influence, with chat and instruct variants consistently exhibiting higher Flesch ease without corresponding gains in empathy or formality.

\begin{figure}[!t]
\centering
\includegraphics[width=\linewidth]{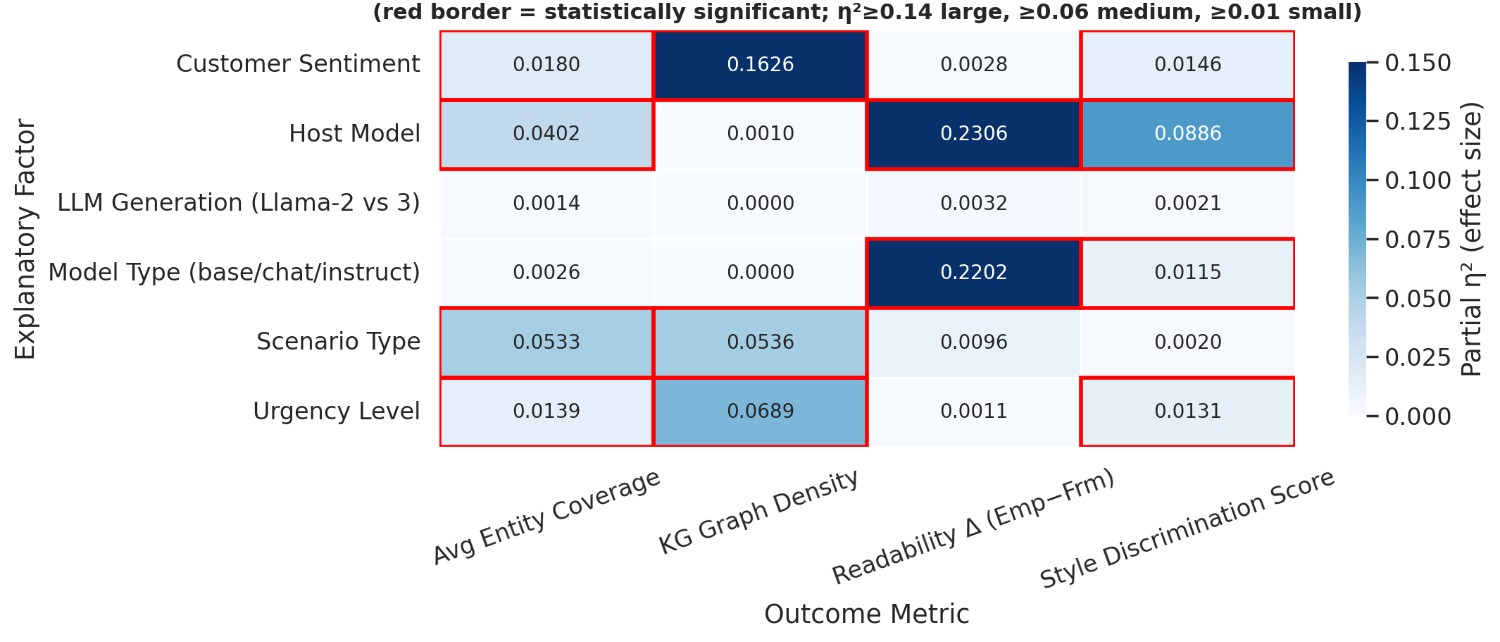}
\caption{Partial $\eta^2$ variance decomposition for each outcome metric using one-way ANOVA. Red borders indicate statistically significant factors.}
\label{fig:eta2}
\end{figure}

This observation is supported quantitatively by the one-way ANOVA variance decomposition in Fig.~\ref{fig:eta2}. Scenario Type ($\eta^2=0.053$) explains more entity-coverage variance than Host Model ($\eta^2=0.040$), indicating that input conditions contribute more strongly to factual grounding than model identity. Collectively, input-condition factors explain more variation in factual grounding than host-model identity.

\begin{figure*}[!t]
\centering
\includegraphics[width=0.85\textwidth]{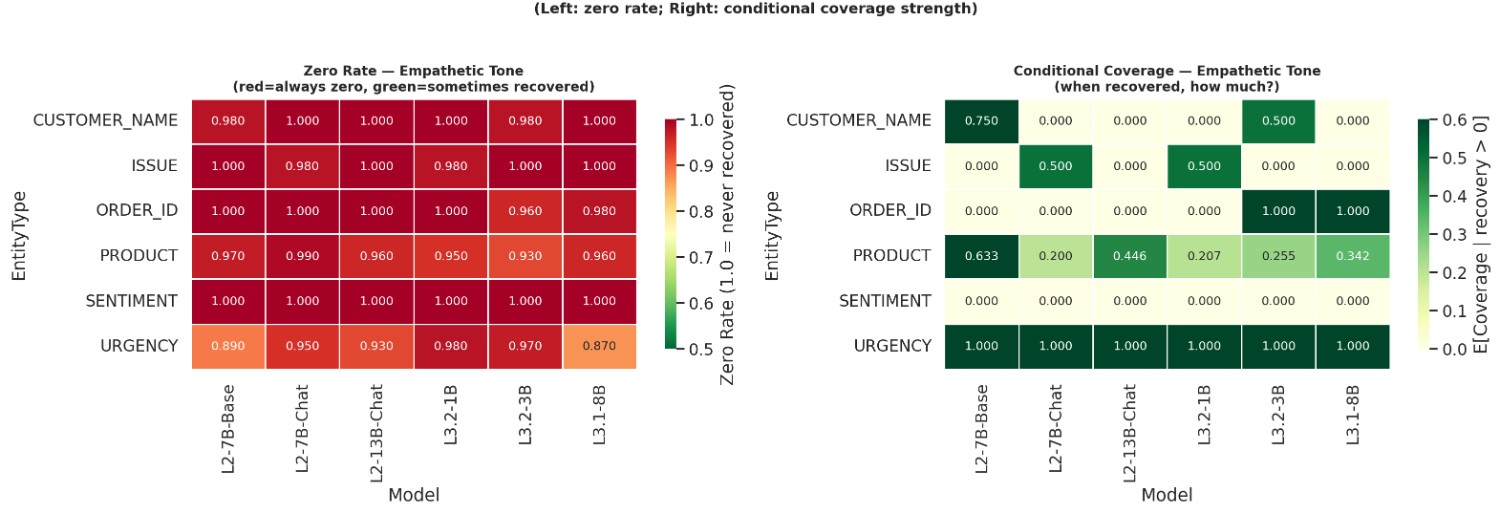}
\caption{Zero-inflation analysis of per-entity-type recovery. Left: zero rate (fraction of cases with no recovery). Right: conditional coverage given at least one recovery.}
\label{fig:zeroinflation}
\end{figure*}

A zero-inflation analysis further exhibits a structural limitation of the shallow ontology, summarized in Fig.~\ref{fig:zeroinflation}: \textsc{Sentiment} is never recovered in the empathetic pipeline, whereas \textsc{Urgency} is the most recoverable entity type. This behavior is consistent with the ontology design: urgency is represented by salient lexical tokens that are readily reproduced, whereas sentiment labels remain entangled with the steered stylistic register.

\subsection{Layer Separability and Steering-Strength Sensitivity}
\label{sec:experiments-h}

\begin{figure*}[!t]
\centering
\includegraphics[width=0.7\textwidth]{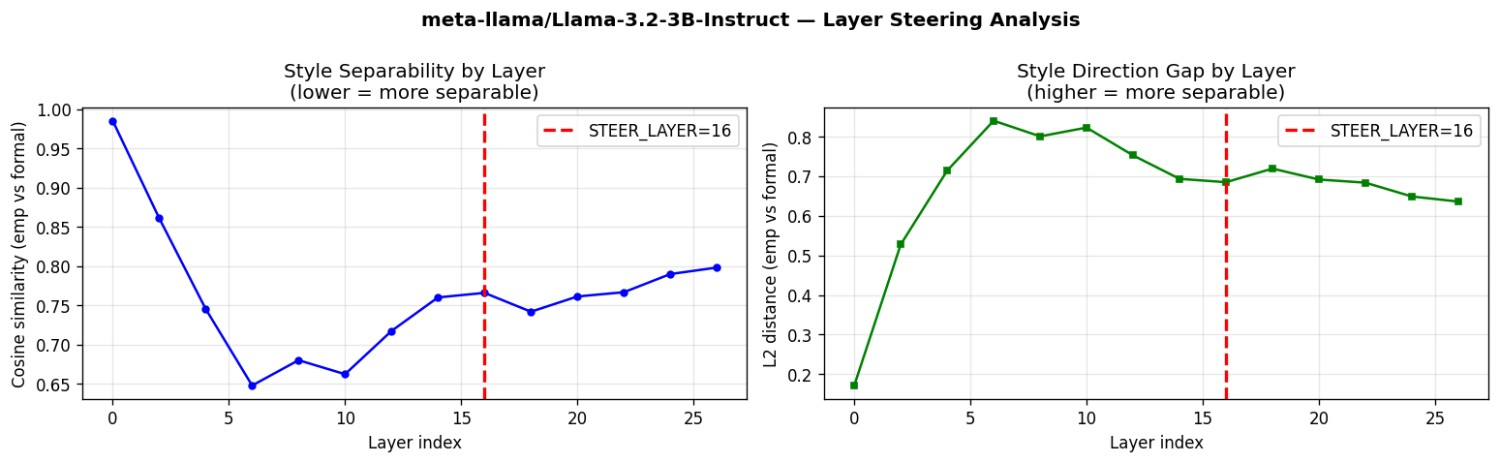}
\caption{Layer-wise separability of empathetic and formal steering directions for LLaMA-3.2-3B-Instruct.}
\label{fig:layer-sep}
\end{figure*}

Two diagnostics characterize the steering operator independently of the KG layer. First, a layer-wise separability probe computes the cosine similarity and $L_2$ distance between empathetic and formal steering directions across transformer layers. For LLaMA-3.2-3B-Instruct, maximum separability occurs around layers 6-10, whereas the deployed layer ($L=16$) lies beyond this optimum, as shown in Fig.~\ref{fig:layer-sep}. This indicates that the alpha-sweep protocol does not necessarily identify the layer with the greatest style separability.

\begin{figure*}[!t]
\centering
\includegraphics[width=0.7\textwidth]{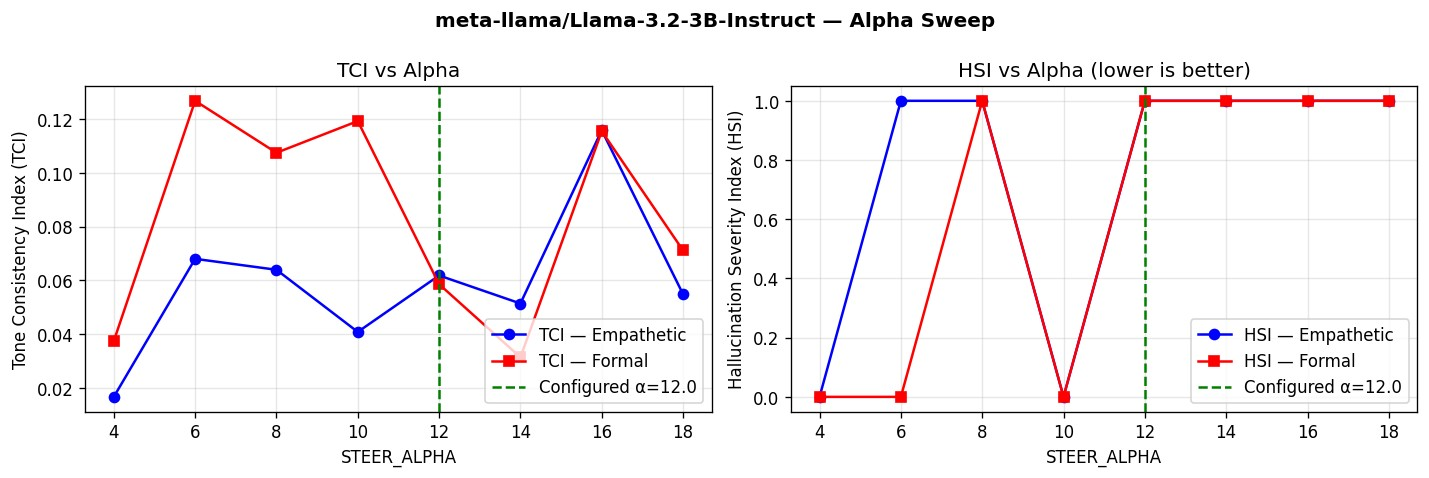}
\caption{Steering-strength sweep for LLaMA-3.2-3B-Instruct at the deployed layer ($L=16$).}
\label{fig:alpha-sweep}
\end{figure*}

Second, steering strength ($\alpha$) is swept at the deployed layer while measuring the TCI (Eq.~\ref{eq:tci}) and HSI (Eq.~\ref{eq:hsi}). As shown in Fig.~\ref{fig:alpha-sweep}, TCI varies non-monotonically with $\alpha$, whereas HSI rapidly saturates over much of the sweep. This behavior appears in two of the six evaluated models (LLaMA-3.2-3B and LLaMA-2-13B-chat), revealing a previously undocumented operating region in which hallucination diagnostics become insensitive to steering strength while tone consistency continues to vary. Because the KG remains fixed throughout the sweep, this behavior is attributable to the steering operator rather than the knowledge-grounding framework.

% ------------
% SECTION IX — Discussion  [EDITED: honest single-model footnote + cross-gen sentence — Points 3 + honest answer]
% ------------
\section{Discussion}
\label{sec:discussion}
The central finding is that entity coverage improves significantly while all style metrics remain unchanged, confirming that a symbolic knowledge layer can coexist with representation-level steering without interference. Zero placeholder leakage across 1,200 generations further confirms the robustness of the mask-and-rehydrate protocol (Eqs.~\ref{eq:defactualize}, \ref{eq:rehydrate}), a consequence of rehydration being a deterministic lookup rather than a generative step. The activation-steering path never produced a malformed response across the full corpus, so this robustness pattern indicates the modest coverage observed stems from the steered model omitting placeholders during generation, not from rehydration failing to restore them.

\textbf{Limitations.}
\label{sec:limitations}
The entity-coverage gain, though statistically significant, is modest in magnitude ($d=0.225$), with absolute recovery ranging from low single digits to low tens of percent across models (Table~\ref{tab:styledisc-entcov}); DSR thus improves, but does not guarantee, complete factual grounding. The effect size itself is a single-model (LLaMA-2-7B-chat) estimate: a Llama-2 vs.\ Llama-3 sub-corpus comparison ($n=300$ each) found no significant cross-family difference in entity coverage, readability, or style discrimination ($p>0.17$ in all cases), with only Flesch Ease differing significantly, suggesting a host-family rather than KG-layer effect. Full six-model replication of the ablation remains future work. Finally, DSR assumes all entities requiring preservation are already present in the input, making it complementary to retrieval-augmented methods that introduce external knowledge rather than a substitute for them.

\textbf{Scalability.} KG construction and steering-vector caching add negligible overhead to a single host-model forward pass (Section~\ref{sec:complexity}) and, unlike PEFT methods with per-user adapter costs~\cite{hu2022lora}, require only one cached vector per (style, model) pair. The main open challenge is cross-turn deployment, where the ephemeral KG would need to become a persistent, indexed store for long-term agent memory~\cite{packer2023memgpt}.

\section{Conclusion}
\label{sec:conclusion}
We addressed semantic leakage in activation-steered, fact-sensitive agentic dialogue as a knowledge-engineering problem by introducing a typed, salience-weighted knowledge graph that masks entities before generation and deterministically restores verified values afterward, without modifying the steering vector, layer, or strength. Across six Llama variants and 600 A2A-generated support cases, the KG layer improved entity coverage (Cohen's $d=0.225$, $p_{\text{Bonf}}=1.0\times10^{-4}$) while preserving all measured style-fidelity metrics. Layer-separability and steering-strength analyses further revealed a previously undocumented operating region, observed in two models, where hallucination diagnostics saturate while tone consistency remains non-monotonic. The results demonstrate how structured knowledge representations can complement representation-level control to improve factual reliability without modifying foundation models. Future work will extend the evaluation across all models, incorporate persistent indexed knowledge graphs for long-term agent memory and graph-based retrieval, and explore richer ontological representations to determine whether the observed KG structural properties generalize beyond the current six-type schema.

\section*{Acknowledgment}
Portions of this work benefited from the supervised use of Claude, and Gemini in drafting and editing. We are grateful to the open-source community for tools enabling this research. The authors gratefully acknowledge the institutional support and funding provided for this work. The authors have no competing interests to declare that are relevant to the content of this article.

\bibliographystyle{IEEEtran}

\end{document}